\documentclass[lettersize,journal]{IEEEtran}
\usepackage{amsmath,amsfonts}
\usepackage{algorithmic}
\usepackage{algorithm}
\usepackage{array}
\usepackage[caption=false,font=normalsize,labelfont=sf,textfont=sf]{subfig}
\usepackage{textcomp}
\usepackage{stfloats}
\usepackage{url}
\usepackage{verbatim}
\usepackage{graphicx}
\usepackage{multirow}
\usepackage[style=ieee, sorting=nyt]{biblatex}
\begin{document}

\title{UrbanTwin: Building High-Fidelity Digital Twins for Sim2Real LiDAR Perception and Evaluation}
\author{Muhammad Shahbaz, Shaurya Agarwal (\IEEEmembership{Member,IEEE})

\thanks{This paper was produced by the IEEE Publication Technology Group. They are in Piscataway, NJ.}
\thanks{Manuscript received April 19, 2021; revised August 16, 2021.}
\thanks{Muhammad Shahbaz and Shauray Agarwal are with the Department of Civil, Environmental and Construction Engineering, University of Central Florida, Orlando, FL 32816, USA.}}

\markboth{Journal of \LaTeX\ Class Files,~Vol.~14, No.~8, August~2021}%
{Shell \MakeLowercase{\textit{et al.}}: A Sample Article Using IEEEtran.cls for IEEE Journals}

\IEEEpubid{0000--0000/00\$00.00~\copyright~2021 IEEE}

\maketitle

\begin{abstract}
LiDAR-based perception in intelligent transportation systems (ITS) relies on deep neural networks trained with large-scale labeled datasets. However, creating such datasets is expensive, time-consuming, and labor-intensive, limiting the scalability of perception systems. Sim2Real learning offers a scalable alternative, but its success depends on the simulation’s fidelity to real-world environments, dynamics, and sensors.
This tutorial introduces a reproducible workflow for building high-fidelity digital twins (HiFi DTs) to generate realistic synthetic datasets. We outline practical steps for modeling static geometry, road infrastructure, and dynamic traffic using open-source resources such as satellite imagery, OpenStreetMap, and sensor specifications.
The resulting environments support scalable and cost-effective data generation for robust Sim2Real learning. Using this workflow, we have released three synthetic LiDAR datasets, namely UT LUMPI, UT V2X Real, and UT TUMTraf I, which closely replicate real locations and outperform real-data-trained baselines in perception tasks. This guide enables broader adoption of HiFi DTs in ITS research and deployment.

%
%
\end{abstract}


\section{Introduction}

LiDAR (Light Detection and Ranging) has emerged as a cornerstone sensing technology for intelligent transportation systems (ITS), offering high-precision, lighting-invariant 3D measurements that underpin perception tasks such as object detection, tracking, and semantic or instance segmentation. These perception capabilities are increasingly powered by deep neural networks, whose performance and generalization critically depend on large-scale, high-quality labeled datasets. For 3D object detection, for example, each point cloud frame must be annotated with bounding box information, capturing location, size, and orientation, for every object of interest (e.g., cars, trucks, bicycles). Over the past decade, numerous benchmark datasets have been curated for different LiDAR-based applications, including KITTI \cite{geiger2013vision}, Waymo Open \cite{Sun2020}, and nuScenes \cite{caesar2020nuscenes} for autonomous driving; IPS300+ \cite{wang2021ips300+}, LUMPI \cite{busch2022lumpi}, and TUMTraf-I \cite{zimmer2023tumtraf} for roadside perception; and DAIR-V2X \cite{yu2022dair}, V2X-Real \cite{xiang2024v2x}, and BAAI-VANJEE \cite{yongqiang2021baaivanjee} for collaborative perception. While these datasets have accelerated research, their creation and expansion face persistent challenges: (1) the high cost, in time, labor, and resources, of recording and annotating large-scale point clouds; (2) annotation errors or biases introduced by human labelers; (3) limited geographic transferability due to variations in traffic composition, environmental conditions, and infrastructure; (4) the scarcity of rare or safety-critical events such as accidents; and (5) shifts in data distribution caused by changes in sensor specifications or installation settings. These constraints limit the scalability and adaptability of LiDAR datasets, creating a pressing need for cost-effective, rapid, and flexible approaches to generate high-quality datasets that closely match real-world distributions while supporting scenario-specific customization and sensor variation.

In response to this challenge, it is presented that simulations leveraging high-fidelity digital twins (HiFi DTs) offer a powerful and scalable alternative through simulation-to-real (Sim2Real) learning. HiFi DTs, in this context, are virtual, geospatially grounded replicas of real-world environments. Unlike traditional synthetic datasets generated from arbitrary or game-like environments, HiFi DTs are anchored in actual urban geometry, incorporating real road layouts, building structures, and elevation models sourced from publicly available geospatial data. Their controlled simulation environment allows precise placement of LiDAR sensors, fine-tuned modeling of occlusions, and seamless integration of dynamic traffic agents, all while maintaining biased-free ground-truth annotations. This level of fidelity and controllability not only enables repeatable experiments and diverse scenario generation but also supports the production of semantically rich, noise-free datasets that align more closely to deployment conditions. 

However, creating HiFi DTs itself is a challenging task as it requires faithfully replicating the structural, spatial, and contextual details of real-world environments. A high-fidelity digital twin must accurately reconstruct the geometry of static elements such as buildings, road surfaces, curbs, trees, etc., preserving their true dimensions, spatial arrangement, and relative positioning within the scene. It must also support the placement of LiDAR sensors in realistic configurations and generate LiDAR returns. This level of realism demands careful integration of diverse geospatial data sources, consistency in coordinate reference systems, and representation of all static and dynamic elements in the simulation environment. In the absence of these components, the resulting digital twin risks oversimplifying the complexities of real-world scenarios, thereby undermining its effectiveness for Sim2Real learning.

This paper addresses precisely this gap. We propose and detail a structured, reproducible methodology to construct high-fidelity digital twins using commonly accessible public geospatial data sources, such as satellite imagery and OpenStreetMap (OSM) data. By integrating these sources, distribution-aware traffic generation, and specification-precise lidar sensor modeling in CARLA \cite{carla}, an established simulation environment, highly faithful point-cloud datasets are created. Such datasets not only resolve limitations discussed above but significantly expand the possibilities for developing LiDAR-based perception models.

Please note that the primary objective of this paper is to present the detailed tutorial/methodology on how to create HiFi DTs and generate synthetic data for lidar-based perception applications. The details on resulting datasets are established in \cite{shahbaz2025urbantwinhighfidelitysyntheticreplicas} where this methodology is used to create and publish three synthetic LiDAR datasetes, UT-LUMPI \cite{ucf_ut_lumpi}, UT-V2X-Real-IC \cite{ucf_ut_v2x_real_ic}, and UT-TUMTraf-I \cite{ucf_ut_tumtraf_i}, replicating three real LiDAR datasets, LUMPI \cite{busch2022lumpi}, V2X-Real \cite{xiang2024v2x}, and TUMTraf-I \cite{zimmer2023tumtraf}, respectively. The efficacy of the approach, in terms of training deep learning models is studied, in-depth, in \cite{shahbaz2025highfidelitydigitaltwinsbridging}
by rigorous feature-level analysis and systematic model evaluation.

The contributions of this paper are:
\begin{itemize}
    \item It develops a feasible and systematic method to create high-fidelity digital twins of real locations using publicly available information.
    \item It presents a technique to accurately model lidar-sensor simulation on top of HiFi DTs.
\end{itemize}

\section{Literature Review}

\noindent\textbf{Limitations of Existing LiDAR Datasets and Simulation Environments:} The development of LiDAR-based perception systems traditionally relies on real and human-labeled point-cloud datasets \cite{zhao2021epointda, Unali24}. While these datasets offer valuable benchmarks, they are often constrained in terms of geographic diversity, annotation cost, and coverage of rare or edge-case scenarios. Simulation environments offer a scalable alternative for dataset generation, providing control over environmental variables and the ability to simulate diverse conditions at negligible cost \cite{li2024choose}. Popular open-source simulators, such as CARLA \cite{dosovitskiy2017carla}, DeepDrive \cite{team2020deepdrive}, and Vista \cite{amini2022vista}, support a range of tasks for perception, planning, and control in autonomous driving systems. Their use has also been extended to support roadside and infrastructure-based perception tasks. However, these platforms typically rely on hand-crafted assets and simplified physics models, leading to synthetic LiDAR data that lacks realism and diverges significantly from real-world point cloud distributions \cite{manivasagam2020lidarsim}. This mismatch creates a pronounced simulation-to-reality (Sim2Real) domain gap, where models trained exclusively on synthetic data perform poorly when deployed in real-world scenarios \cite{8864642, nowruzi2019much}.

\begin{figure*}[!h!t]
\centerline{\includegraphics[width=1\textwidth]{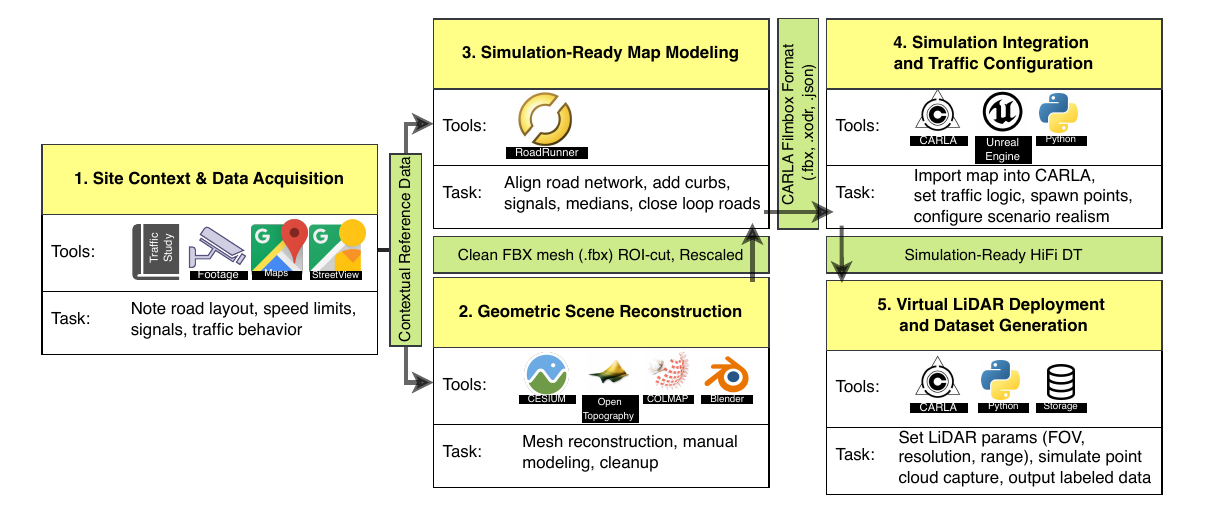}}
\caption{An overview of the Framework for High-Fidelity Digital-Twin Modeling
\label{fig:framework_for_hifi_dt}}
\end{figure*}

\vspace{0.5em}
\noindent\textbf{Assessing Dataset Quality: Realism, Distribution, and Diversity:} The usefulness of a simulation generated dataset extends beyond its volume; it depends critically on three core attributes: realism, statistical distribution, and environmental diversity \cite{nowruzi2019much}. Realism depends on the fidelity of 3D geometry, material properties, and sensor emulation. Distribution refers to whether the generated data reflects the frequency and co-occurrence patterns of real-world traffic scenarios. Diversity concerns coverage of object types, densities, environmental layouts, and dynamic interactions. Many existing simulators fall short on these dimensions due to asset reuse, uniform scene layouts, or lack of fine-grained geographic grounding. As a result, synthetic datasets often fail to capture rare but operationally critical scenarios, limiting their value for training robust models.

\vspace{0.5em}
\noindent\textbf{Emergence of Digital Twin-Based Simulation:} To overcome the Sim2Real gap, recent work has explored using digital twins, high-fidelity, geospatially accurate virtual replicas of real-world environments, to synthesize more realistic and domain-aligned datasets. For example, Li et al. \cite{li2025digital} constructed a digital twin-based environment to generate labeled image datasets and applied graph-based domain alignment to bridge feature-level gaps between synthetic and real data. Their model, which combined multi-task learning with domain adaptation via discriminators, improved transferability across domains. Similarly, Strunz et al. \cite{strunz2024cross} introduced the LUCID dataset, a synthetic LiDAR dataset explicitly modeled to mirror the spatial and distributional characteristics of real-world datasets like KITTI and nuScenes. Their cross-dataset evaluation demonstrated that models trained on LUCID achieved superior generalization, with smaller deltas in average precision and recall compared to models trained on other synthetic or even real datasets.

These studies underscore the promise of digital twins for improving dataset fidelity and reducing the domain gap. However, building high-quality digital twins is a complex and resource-intensive process. It requires accurate modeling of road geometries, infrastructure elements, elevation profiles, and dynamic agents, all while maintaining geospatial accuracy and sensor realism. The construction of such environments is further complicated by the need to integrate heterogeneous data sources (e.g., satellite imagery, OpenStreetMap, and elevation data) and harmonize them into simulation-ready formats. These challenges motivate the need for pipelines that can transform publicly available geospatial data into high-fidelity simulation environments suitable for LiDAR-based Sim2Real learning. In the next section, we present our step-by-step tutorial/methodology to create HiFi DTs of real locations and generating LiDAR data using them. The generated LiDAR data has been comprehensively tested in \cite{shahbaz2025highfidelitydigitaltwinsbridging} proving the efficacy of the approach.

\section{Methodology}

\begin{figure*}[!hb]
\centerline{\includegraphics[width=0.98\linewidth]{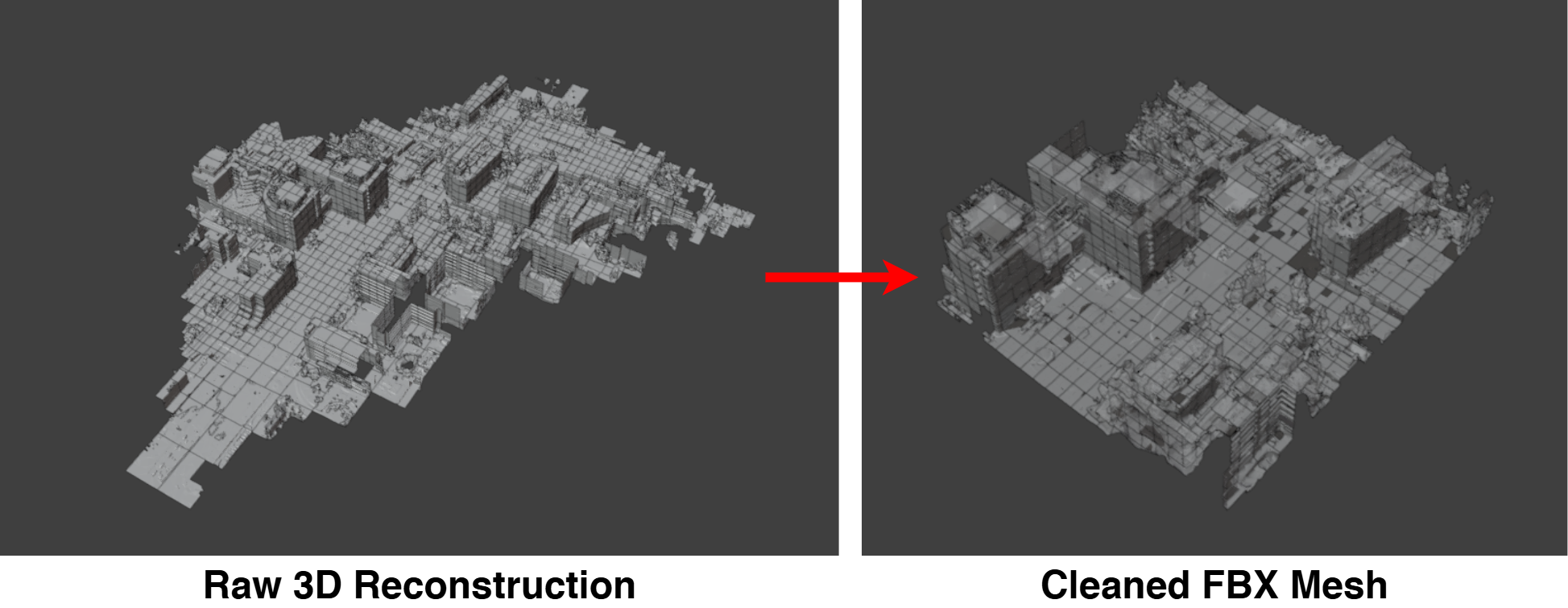}}
\caption[Cleaning Raw 3D Geometry Reconstruction]{\textbf{Left: }A raw reconstruction of 3D mesh of real location. \textbf{Right: }The reconstruction is cleaned in blender to region of interest.
\label{fig:reconstruct_to_clean}}
\end{figure*}

Our method for constructing high-fidelity digital twins (HiFi DTs) for LiDAR-based perception comprises five structured stages. Each stage incorporates specific tools and best practices for constructing HiFi DTs. An overview of the stages is presented in Fig.~\ref{fig:framework_for_hifi_dt}.

\subsection{Site Analysis and Contextual Data Acquisition}

\begin{figure*}[!h]
\centerline{\includegraphics[width=0.98\linewidth]{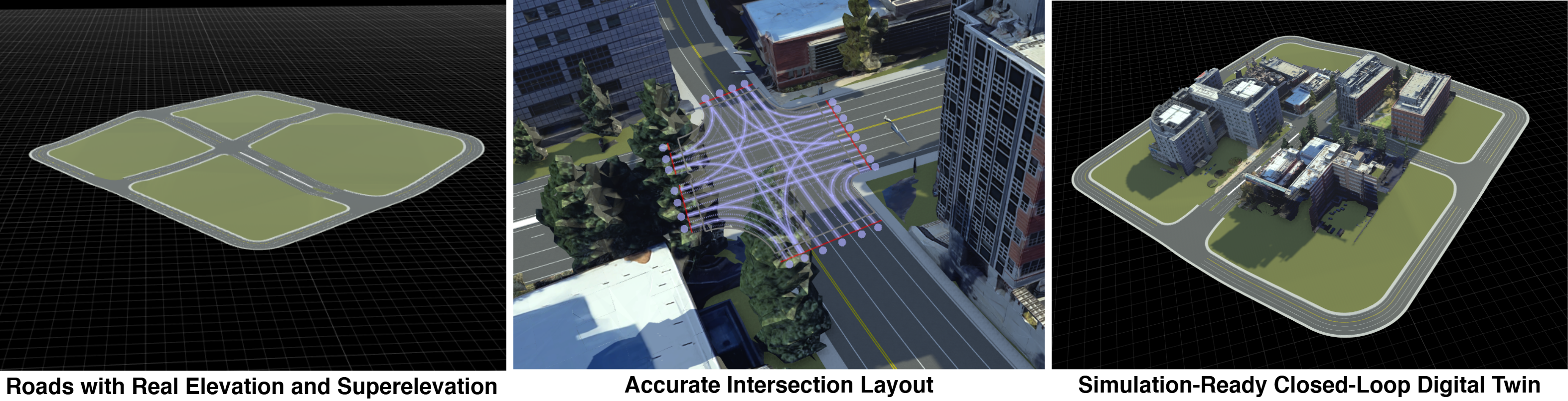}}
\caption[Roadnetwork Modeling in RoadRunner]{\textbf{Left: }The sub-figure shows that roads are designed in Roadrunner constrained on lane-level elevation and superelevation. \textbf{Middle: } Special care is taken in modeling road network and traffic maneuvers at intersections. \textbf{Right: } The digital-twin is made close loop so traffic distribution remains fixed in the region. 
\label{fig:sim_ready}}
\end{figure*}

\begin{figure*}[!h!b]
\centerline{\includegraphics[width=0.98\linewidth]{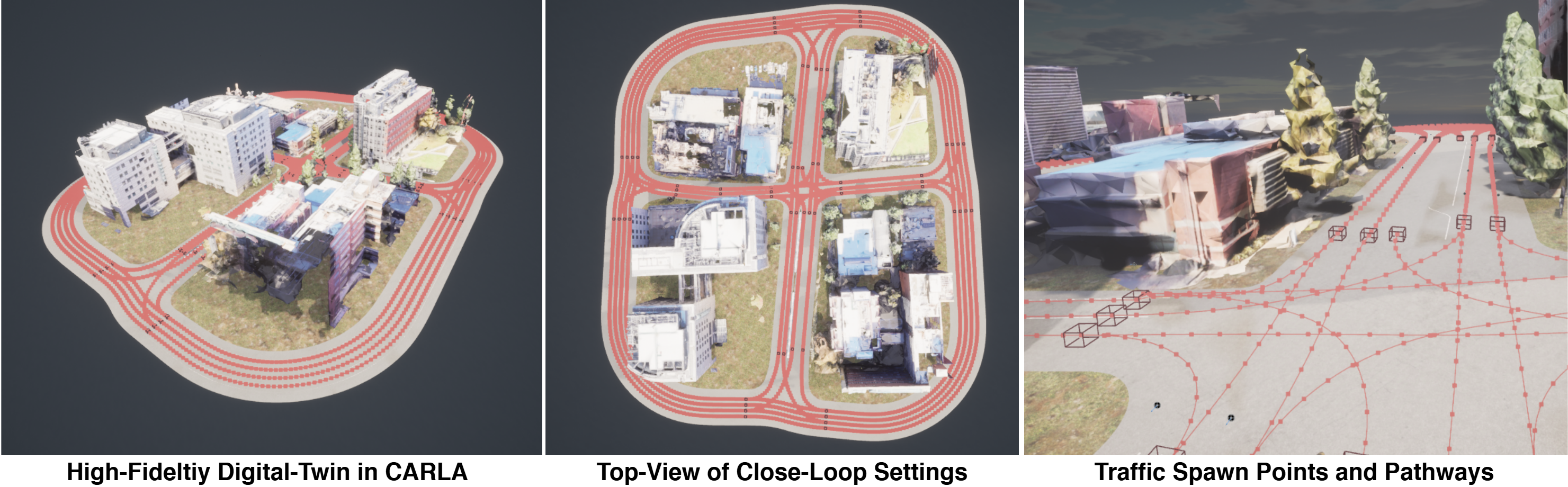}}
\caption[Digital Twin - Visuals]{\textbf{Left: }A perspective view of HiFi DT in CARLA. \textbf{Middle: } Top view of the same DT. The red lines are the traffic paths. The DT is constrcuted in loop to restrain the actor distribution in the simulation. \textbf{Right: } A close-up on how traffic maneuvers are constructed at intersection. The small black spheres show traffic spawn points. 
\label{fig:dt_vis}}
\end{figure*}

The process begins with identifying the target geographic location and collecting contextual information about the site. Using publicly available satellite imagery and street-level views, such as via Google Maps \cite{google_maps}, high-level information is collected, including road layout, speed limits, lane configurations, traffic signals, and surrounding structures. This information can be supplemented by traffic surveillance footage or publicly available transportation studies to better understand traffic composition and flow. The purpose of this stage is to gather contextual reference data for later stages, ensuring accurate structure, traffic behavior, and environment semantics.

\subsection{Geometric Scene Reconstruction}

To replicate the static environment, a 3D reference model of the region of interest (ROI) is required. This model can be sourced from platforms offering 3D tiles or topographical data, such as Cesium \cite{cesium_ion}, OpenTopography \cite{opentopography}, or municipal GIS portals. If a ready-made model is unavailable or lacks resolution, the scene may be reconstructed using photogrammetry tools, such as Autodesk ReCap Pro \cite{autodesk_recappro}, RealityCapture \cite{realitycapture}, or COLMAP \cite{colmap}. If 3D modeling expertise are available, then static environment can also be modeled manually using 3D modeling software such as Autodesk Maya \cite{autodesk_maya}, Cinema 4D \cite{cinema4d}, or Blender \cite{blender}, however, reference 2D images should be used to model geometry to align it accurately to real-world.

If the 3D scene is not manually modeled, it often needs some preprocessing, that can also be done in 3D modeling software. It is because the model is an estimation of the structures in the target reality and contains low-poly geometries. However, since real lidar data is also inherently sparse, therefore low-level surface details are not important in this use case. Another limitation of a reconstructed 3D model is that, often times it can contain traffic blobs because the source 2D images, that were used to reconstruct mesh, had such noise present. Another source of noise can be the reconstruction algorithm itself as it not perfect and can introduce noise like adding extra geometry that is not present in the real scene. Moreover, some geometry data outside the region of interest can also be still present in the captured 3D image and needs removal for efficiency reasons. Fortunately, removing such noise is rather easy using any 3D modeling tool such as Blender. By loading the entire 3D mesh in 3D modeling tool, the mesh can be cut to region of interest, followed by removing extra geometry (traffic blobs, 3D shards in mid air, etc.) based on qualitative inspection. Additionally, the model is rescaled to match real-world units. Fig.~\ref{fig:reconstruct_to_clean} shows before and after pre-processing of a raw 3D reconstruction. Optionally, for convenience later, sensor positions are also noted at this stage. The resulting scene is exported in the Filmbox (FBX) format, that is compatible with road editing tools.

\subsection{Simulation-Ready Map Modeling}

The FBX file is imported into RoadRunner \cite{mathworks_roadrunner}, a specialized road network design tool with native support for CARLA. Within RoadRunner, road geometries are aligned with the 3D reference model to ensure scale consistency. Lane widths are calibrated using known defaults, and elevation and superelevation profiles are adjusted to match observed terrain.

Roads beyond the ROI are looped to create a closed topology, allowing continuous vehicle movement without requiring real-time respawning. Curbs, medians, etc. are manually added using \texttt{Traffic Island Tool} to match the real-world scene. Signal logic and speed limits are set using \texttt{Signal Tool} and set to the match observations in initial site analysis. Fig.~\ref{fig:sim_ready} shows simulation ready HiFi DT developed in RoadRunner.

\subsection*{Simulation Integration and Traffic Configuration}

\begin{figure*}[!h!t]
\centerline{\includegraphics[width=0.98\linewidth]{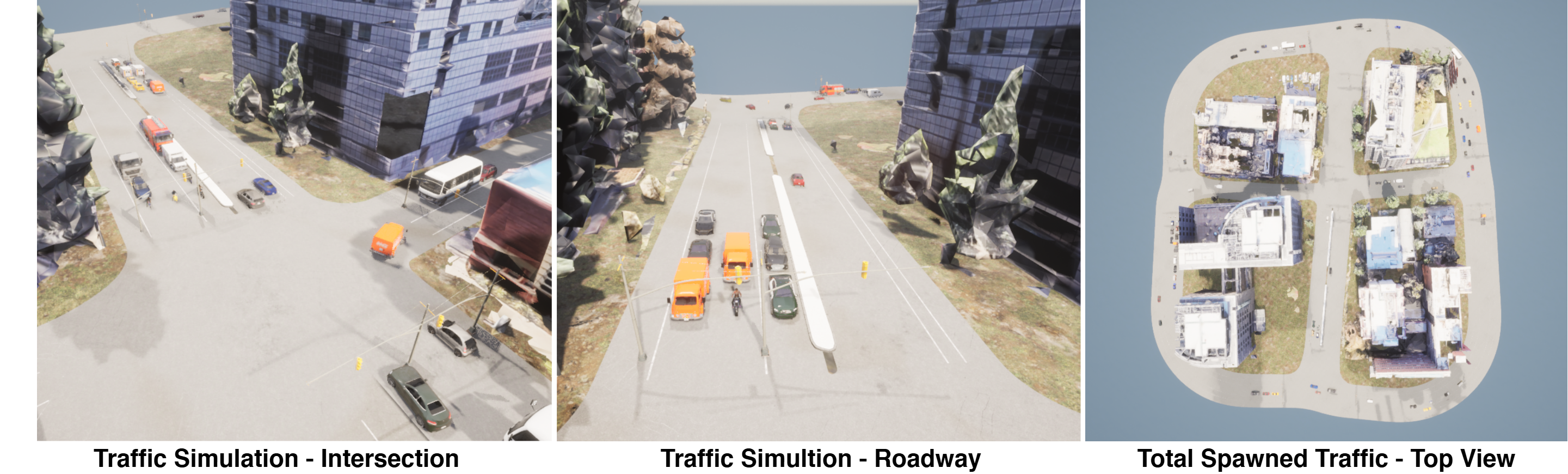}}
\caption[Digital-Twin Simulation in Action]{Snapshots of simulation in action.\textbf{Left: }The vehicles at intersection. It can be observed that traffic is following signal routine created in RoadRunner. \textbf{Middle: } A scene from the road showing mixed traffic (bikes, cars, trucks). \textbf{Right: } Top view of entire traffic scene. The loop in the DT makes it easier to spawn the traffic once and then record data without respawning. 
\label{fig:dt_in_action}}
\end{figure*}

Once the road network is complete, the map is exported from RoadRunner in \textit{CARLA FilmBox} format and integrated into CARLA using the official import pipeline. The \texttt{Import.py} script, located in the \texttt{Util/BuildTools} directory of the CARLA (build-from-source version), is used to compile the new map into a usable CARLA environment. Fig.~\ref{fig:dt_vis} shows a HiFi DT imported into CARLA environment. Additional documentation for building CARLA from source is available at: \url{https://carla.readthedocs.io/en/latest/build_carla}. It is recommended to use built-from-source version of CARLA as it supports editing the final simulation map directly inside Unreal Engine \cite{unreal_engine}.

After importing, traffic spawn points are checked and duplicated as needed within the Unreal Engine editor to ensure compatibility to required traffic density and distribution. This is basically counting spawn points and see if the total spawn points are equal or more than total traffic objects (cars, trucks, bikes, etc.) required for simulation. The total number of traffic objects can be a rough estimate of the reality (observed during first stage). This is because stochasticity of traffic spawning processes actually help create more generalized distribution and exact count is therefore not needed. CARLA provides a \texttt{generate\_traffic.py} script that is highly adjustable using command-line arguments. Once the simulator (CARLA in Unreal Engine) is running, this script can be used to simulate dynamic agents. It allows selection and control over the type and distribution of vehicles to reflect realistic traffic conditions. For instance, if real-world traffic exhibits a high proportion of passenger cars relative to buses, the script can bias spawn probabilities accordingly. Users may also exclude uncommon vehicle types or introduce custom models, depending on the use case (e.g., work zone simulations or emergency scenarios).

Before full-scale data generation, a brief interactive simulation is run to validate traffic behavior and identify potential anomalies, such as traffic jams due to misaligned roads or spawn point conflicts. Fig.~\ref{fig:dt_in_action} show the traffic simulation running in CARLA. 

\begin{figure*}[!h!b]
\centerline{\includegraphics[width=0.98\linewidth]{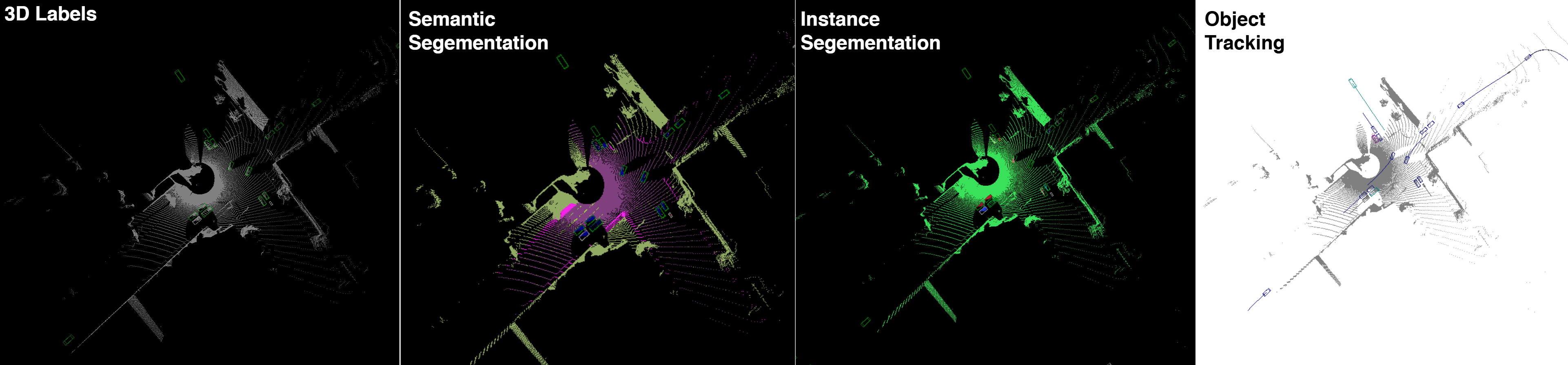}}
\caption[Visualization of Supported Label Types in HiFi-DT-based Datasets]{Label types (3D bounding box, semantic segmentation, instance segmentation, and object tracking) visualized.\textbf{Left Top: }3D bounding box labels; even empty bounding boxes outside virtual LiDAR range are also imported to support occlusion and tracking applications. \textbf{Top Right: } Semantic segmentation; labels includes background, road, and foreground. \textbf{Left Bottom: } Object-level segmentation are supported to train models to segment individual points. \textbf{Right Bottom: }The output labels also include tracking for each object to support tracking and prediction application for LiDAR-based ITS perception. 
\label{fig:label_types}}
\end{figure*}

\subsection*{Virtual LiDAR Deployment and Dataset Generation}

To generate training data, virtual LiDAR sensors are deployed within the simulation environment using CARLA's Python API. CARLA provides numerous examples under \texttt{PythonAPI/examples} folder to help users understand sensor spawning process. Moreover, in-depth documentation for spawning and configuring sensors is available at: https://carla.readthedocs.io/en/latest/core\_sensors/. It is recommended to create a single Python script to automate sensor placement and data gather operation. We provide one example script at our Github repository: https://github.com/m-shahbaz-kharal/HiFi-DT-Utils/blob/main/set\_sensors\_and\_gen\_data.py.

The goal of configuring virtual sensors is to replicate real-world hardware specifications, including parameters such as number of channels, horizontal resolution, horizontal and vertical fields of view, measurement range, and point rate. The sensors' specifications also need to be converted to simulation scale in order to generate LiDAR data matching real-world scale. For example, simulations in CARLA have 100 units to 1m ratio, that means 100 units of length in CARLA equates to 1 meter in real world. Such scales should be considered carefully during sensor configuration. 

It is also crucial that sensors are placed at virtual locations that match to real locations in real world. This can be achieved using available reference data or heuristic cues such as pavement markings, infrastructure alignment, or intersection design. Once located on the map, the pose information (height, tilt, etc.) is used to fine-tune placement. Our example script, link provided above, is designed to be modular and generalizable, and supports flexible configuration of sensors (both specification and placement) and dataset format.

Upon execution, the simulation environment captures LiDAR point clouds along with associated ground-truth annotations. The outputs include: (a) 3D bounding boxes with object class labels, (b) semantic segmentation masks, (c) instance-level segmentation labels, and (d) unique object IDs for temporal tracking. The supported labels are shown in Fig.~\ref{fig:label_types} using LiGuard's \cite{shahbaz2025} visualizer. The resulting dataset is exported in a modified version of the OpenPCDet format \cite{od2020openpcdet} for compatibility with standard LiDAR perception pipelines.

Intuitively, due to digital-twin modeling, the resulting simulation closely approximates the reality, as the 3D spatial representation and traffic dynamics exhibit a high degree of correspondence to those observed in the real world. This approach offers a cost-effective and computationally efficient approximation, providing sufficient fidelity for the intended applications. Please note that HiFi DT's specifications can either be aligned to existing data for augmentation purposes or customized to match specific experimental scenarios to develop risk-free extreme case lidar datasets.

\begin{figure*}[!t]
\centering
\subfloat[Hausdorff Distance (P95)]{\includegraphics[width=0.32\linewidth]{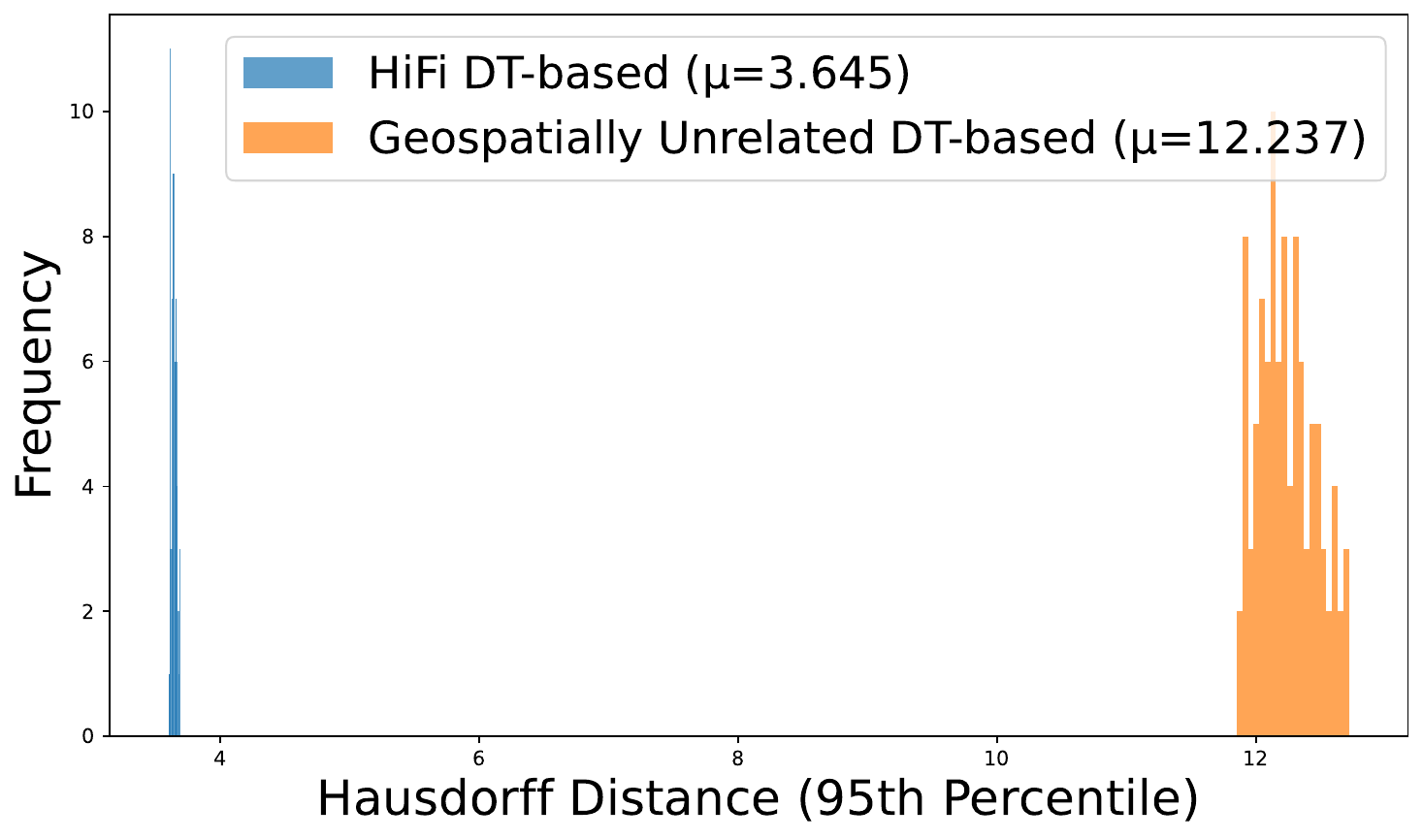}}
\hfill
\subfloat[JS Divergence]{\includegraphics[width=0.32\linewidth]{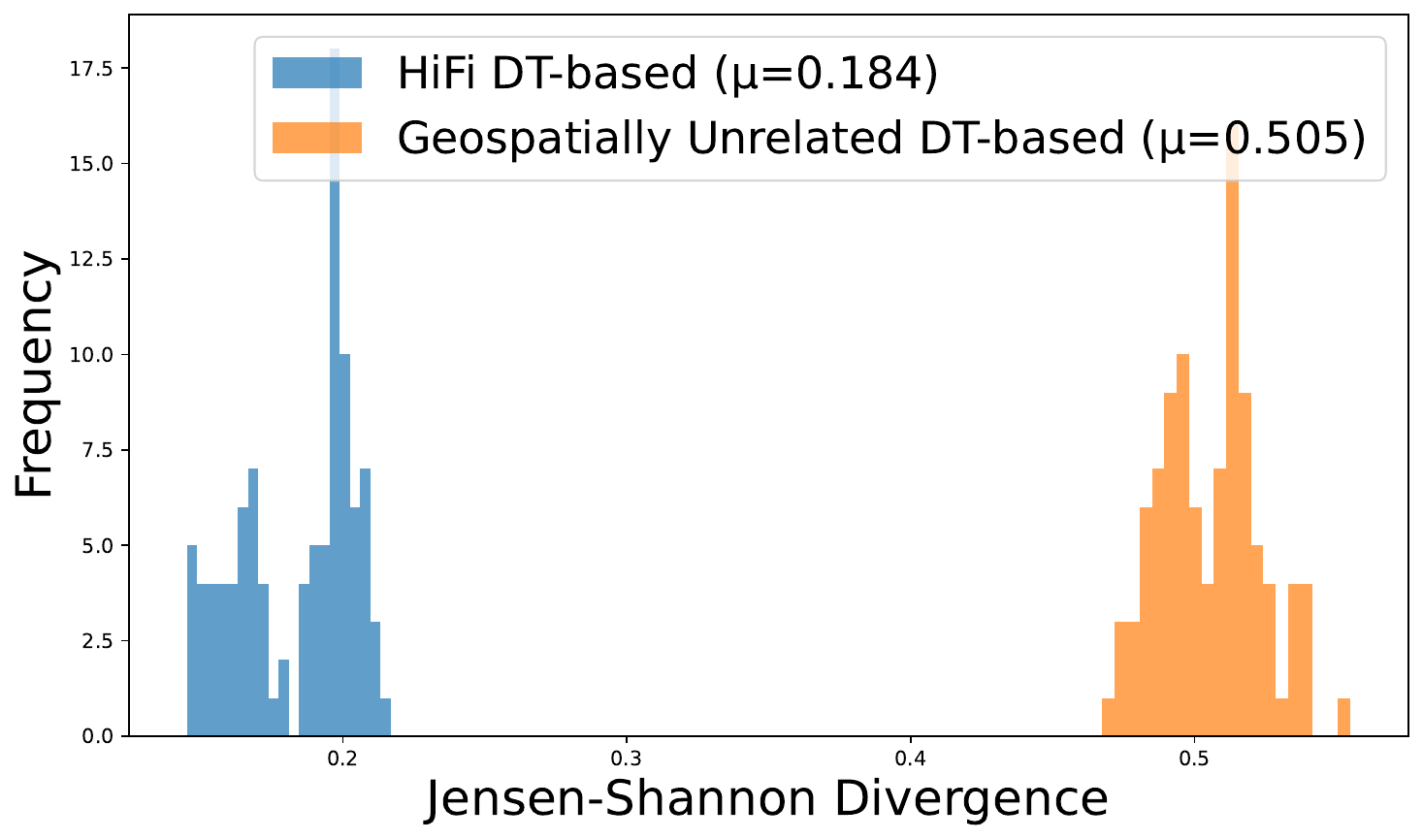}}
\hfill
\subfloat[Point-to-Mesh Distance]{\includegraphics[width=0.32\linewidth]{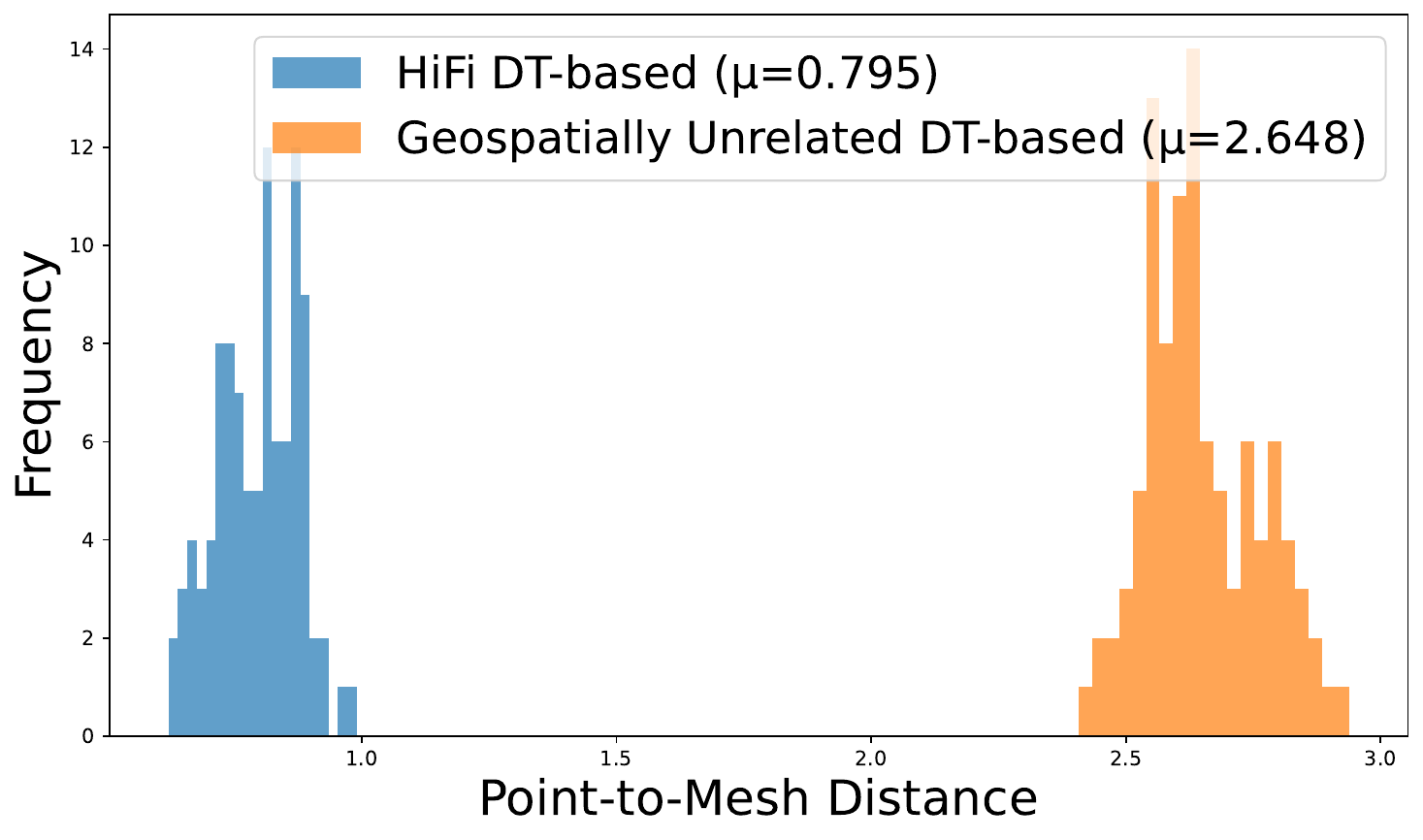}}
\caption{Distributions of similarity metrics comparing Digital Twin (DT) synthetic data to conventional synthetic data created with the same toolchain but for an arbitrary location. Lower values indicate closer resemblance to real data.}
\label{fig:metric_histograms}
\end{figure*}

\begin{figure}[!b]
\centering
\includegraphics[width=0.95\linewidth]{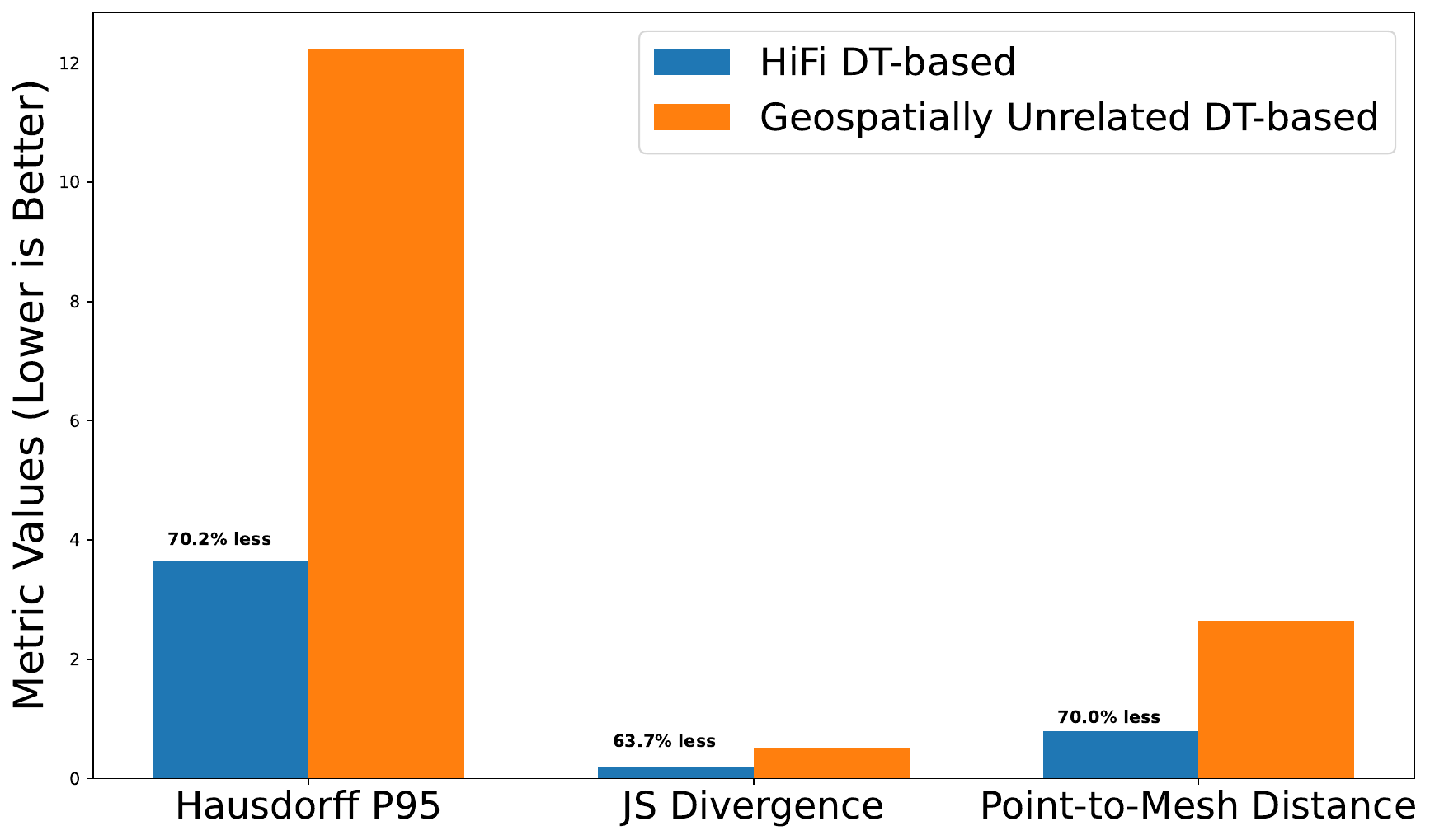}
\caption{Aggregate comparison (lower is better). DT outperforms the arbitrary scene across all metrics: P95 Hausdorff (\textbf{--70.2\%}), JS Divergence (\textbf{--63.7\%}), and P2M (\textbf{--69.9\%}).}
\label{fig:metrics_summary}
\end{figure}

\section{Quantitative Evaluation of Digital Twin Fidelity}
\label{sec:dt_eval}

Building on the pipeline described above, we created two synthetic datasets: (i) a dataset generated using High-Fidleity Digital Twin (HiFi DT) that geospatially replicates the target site, and (ii) a geospatially unrelated synthetic dataset created for a different (arbitrary) location. Crucially, the simulator (CARLA), map-authoring workflow (RoadRunner), traffic scripting, and LiDAR sensor emulation were held constant across both to for fairness. This design isolates the effect of geospatial grounding, that is, if all synthetic data were equally useful, then the arbitrary scene should match real data as well as the HiFi DT. Our hypothesis is that only the HiFi DT-based data preserves the target site’s geometry and usage patterns, produces distributions that align with reality.

We assess realism using three complementary metrics, each capturing a different failure mode of synthetic data, and report their distributions in Fig.~\ref{fig:metric_histograms}(a--c). Aggregate summaries appear in Fig.~\ref{fig:metrics_summary}. We intentionally omit the raw (max) Hausdorff distance due to its sensitivity to isolated outliers; the 95th percentile is more diagnostic of overall structural alignment for LiDAR point sets.

\noindent \textbf{Hausdorff Distance (P95):} The P95 of the bidirectional Hausdorff distance characterizes the typical worst-case separation between two point sets after discounting extreme outliers. Lower is better, indicating that most of the synthetic points lie close to their real counterparts. As shown in Fig.~\ref{fig:metric_histograms}(a), the DT sharply shifts the distribution leftward, reducing P95 by \textbf{70.2\%} relative to the arbitrary scene (means: \textbf{3.645} vs.\ 12.237). This indicates that the DT recreates the large-scale layout (road surfaces, facades, curbs) that governs LiDAR returns, not merely isolated details.

\noindent \textbf{Jensen--Shannon (JS) Divergence:} JS Divergence measures the distance between probability distributions (here, point-level spatial/feature distributions). It answers whether the frequency and co-occurrence patterns in synthetic data resemble those in reality (e.g., where points tend to cluster, how density varies by range). Lower is better. In Fig.~\ref{fig:metric_histograms}(b), the DT achieves a \textbf{63.7\%} reduction (means: \textbf{0.184} vs.\ 0.505), showing that DT-based simulation reproduces scene statistics, such as traffic mix, occlusions, background/foreground balance, far more faithfully than an arbitrary scene built with the same tools.

\noindent \textbf{Point-to-Mesh (P2M) Distance:} P2M evaluates geometric faithfulness by measuring how far synthetic LiDAR points lie from reconstructed real-world surfaces. Lower values mean the simulated returns “sit” on the same physical structures as in reality. Fig.~\ref{fig:metric_histograms}(c) shows a \textbf{69.9\%} reduction for the DT (means: \textbf{0.795} vs.\ 2.648), indicating tight geometric concordance with real surfaces across the site.

\noindent \textbf{Takeaway:} Across structural (P95), statistical (JS), and geometric (P2M) views, DT-based data consistently attains lower (better) values than conventional synthetic scenes generated with identical software stacks. This means that the advantage is not the simulator but the digital twin’s geospatial fidelity to the target location that closes the Sim2Real gap for roadside LiDAR.

These evaluations inform that HiFi DTs show high similarity to real data. This is further tested in terms of statistical and distributional similarity in \cite{shahbaz2025urbantwinhighfidelitysyntheticreplicas} and in terms of effectiveness in training real-world object detection models in \cite{shahbaz2025highfidelitydigitaltwinsbridging} where we show that models trained on synthetic data generated via the methodology discussed in this tutorial paper, perform on par and in some cases exceed baselines trained on real data.

\section{Applications to AI}

Beside generating close-to-real synthetic LiDAR data and labels that can be used in training perception models for real-world systems \cite{shahbaz2025urbantwinhighfidelitysyntheticreplicas, shahbaz2025highfidelitydigitaltwinsbridging}, High-Fidelity Digital Twins (HiFi DTs) enable concrete, repeatable AI workflows that are hard to realize with real-only data. Below we outline applications that are directly supported by the workflow in this tutorial and are readily implementable with standard LiDAR perception stacks.

\noindent\textbf{A. Supervision at Scale without Manual Labels: } Perhaps the most obvious one is that since each simulation pass yields synchronized point clouds with turnkey labels inlcuding 3D boxes, instance IDs, semantic masks, and tracklets (Fig.~\ref{fig:label_types}), joint training for detection, segmentation, and multi-object tracking can be realized without manual annotation. Multiple label types can also be used to stabilize representation learning at training time; at fine-tuning time, only the task(s) of interest need to be retained. Moreover, simulation allows generating data at increasing level of difficulty (traffic density, occlusion level, LiDAR range dropouts, etc.) enabling curriculum learning schedules that start from uncluttered scenes and progress
to dense, occluded intersections.

\noindent\textbf{B. Rare Events Coverage and Stress Testing: } Real-world datasets usually contain commonly occurring scenarios and are limited in scenarios that occur rarely. Since our HiFi DTs are much more aligned to reality, rare classes/events (e.g., bicycles at night, work-zone cones, near-conflict interactions) can be injected by adjusting spawn priors and signal timing. Moreover, it enables targeted synthesis around concrete failure modes (e.g., far-range trucks, partial occlusions, etc.). With fixed random seeds and preserved scenario parameters, these cases form a reusable regression suite where model change, not scene drift, explain performance deltas. Counterfactuals are also straightforward, just replicate a real incident in the DT and vary exactly one factor (sensor height, phase offset, added cyclist) to test causal sensitivity without field intervention.

\noindent\textbf{C. Sensor Layout and Configuration Co-Design: } A key advantage of HiFi DTs is pose and spec search, that is, treat detection/tracking AP as an objective and search over sensor heights, tilts, and vertical FoV/channel counts within the DT. The chosen configuration can then deployed in the field with higher confidence. Similarly, single LiDAR vs. multi LiDAR placements can be tested, with time sync jitter, and partial outages to choose fusion strategies and redundancy levels before hardware purchase. DTs can also be used for testing practical V2X links as they allow inter-sensor baselines and latency injection, enabling clean evaluation of early-/mid-/late-fusion and spatiotemporal alignment methods without confounds from unknown real-world delays. By varying the number of shared proposals/feature maps across infrastructure nodes, one can quantify mAP vs.\ bitrate curves.

\noindent\textbf{D. Generalizing via Weak Supervision and Foundation-Style Representation Learning: } Models can be improved for real data streams by teacher–student training. For example, train a strong teacher on DT; generate pseudo-labels on unlabeled real streams; refine a student with confidence filtering. Periodically refresh the DT with updated traffic mix to limit drift. Generalization in such setups can be improved by using DT to pretrain with strong geometric/photometric perturbations, then enforce temporal/augmentation consistency on real sequences to reduce reliance on dense labels.

Another generalization strategy it to utilize large DT corpora for training foundation models via masked reconstruction or next-sweep prediction pretext tasks to get generic LiDAR encoders; the same backbone transfers to detection, segmentation, and forecasting with thin task heads. By building DTs of several distinct intersections (grid, roundabout, multi-leg arterial) encourages invariances to layout while preserving metric geometry, that can be useful for statewide deployments with varied infrastructure.

\noindent\textbf{E. Privacy-Preserving Data Sharing and Benchmarking: } DT datasets avoid Personally Identifiable Information (PII) that is present in real recorded sequences. Instead DTs has geometry, actor distributions, and sensor specs which can be shared without hesitation of PIIs for standardization.

\section{Conclusion}
This paper presented a structured and reproducible methodology for generating synthetic LiDAR datasets using high-fidelity digital twins (HiFi DTs). By integrating publicly available geospatial data, scalable road modeling, realistic traffic simulation, and virtual LiDAR sensor deployment, the proposed approach enables cost-effective and domain-consistent dataset generation for roadside perception tasks.

The methodology addresses key limitations identified in prior research, particularly the challenges of annotation scalability, domain alignment, and realism. It offers a practical framework that generalizes across geographic locations and sensor configurations. Its modular design supports both the augmentation of existing datasets and the creation of entirely new simulated environments.

The efficacy of this methodology has already been demonstrated across a series of published studies, where HiFi DT–based datasets have been shown to closely replicate real-world distributions and, in some cases, outperform real-data-trained models in LiDAR perception tasks. This tutorial formalizes the creation process, providing step-by-step guidance and to enable broader adoption of HiFi DTs in Intelligent Transportation Systems (ITS) research and deployment.




\printbibliography

@article{geiger2013vision,
  title={Vision meets robotics: The kitti dataset},
  author={Geiger, Andreas and Lenz, Philip and Stiller, Christoph and Urtasun, Raquel},
  journal={The International Journal of Robotics Research},
  volume={32},
  number={11},
  pages={1231--1237},
  year={2013},
  publisher={Sage Publications Sage UK: London, England}
}

@article{wang2021ips300+,
  title={Ips300+: a challenging multimodal dataset for intersection perception system},
  author={Wang, Huanan and Zhang, Xinyu and Li, Jun and Li, Zhiwei and Yang, Lei and Pan, Shuyue and Deng, Yongqiang},
  journal={arXiv preprint arXiv:2106.02781},
  year={2021}
}

@misc{yongqiang2021baaivanjee,
      title={BAAI-VANJEE Roadside Dataset: Towards the Connected Automated Vehicle Highway technologies in Challenging Environments of China}, 
      author={Deng Yongqiang and Wang Dengjiang and Cao Gang and Ma Bing and Guan Xijia and Wang Yajun and Liu Jianchao and Fang Yanming and Li Juanjuan},
      year={2021},
      eprint={2105.14370},
      archivePrefix={arXiv},
      primaryClass={cs.CV}
}

@article{od2020openpcdet,
  title={Openpcdet: An open-source toolbox for 3d object detection from point clouds},
  author={OD Team and others},
  journal={OD Team},
  year={2020}
}

@inproceedings{caesar2020nuscenes,
  title={nuscenes: A multimodal dataset for autonomous driving},
  author={Caesar, Holger and Bankiti, Varun and Lang, Alex H and Vora, Sourabh and Liong, Venice Erin and Xu, Qiang and Krishnan, Anush and Pan, Yu and Baldan, Giancarlo and Beijbom, Oscar},
  booktitle={Proceedings of the IEEE/CVF conference on computer vision and pattern recognition},
  pages={11621--11631},
  year={2020}
}

@inproceedings{yu2022dair,
  title={Dair-v2x: A large-scale dataset for vehicle-infrastructure cooperative 3d object detection},
  author={Yu, Haibao and Luo, Yizhen and Shu, Mao and Huo, Yiyi and Yang, Zebang and Shi, Yifeng and Guo, Zhenglong and Li, Hanyu and Hu, Xing and Yuan, Jirui and others},
  booktitle={Proceedings of the IEEE/CVF Conference on Computer Vision and Pattern Recognition},
  pages={21361--21370},
  year={2022}
}

@inproceedings{busch2022lumpi,
  title={LUMPI: The Leibniz University Multi-Perspective Intersection Dataset},
  author={Busch, Steffen and Koetsier, Christian and Axmann, Jeldrik and Brenner, Claus},
  booktitle={2022 IEEE Intelligent Vehicles Symposium (IV)},
  pages={1127--1134},
  year={2022},
  organization={IEEE}
}

@inproceedings{zimmer2023tumtraf,
  title={Tumtraf intersection dataset: All you need for urban 3d camera-lidar roadside perception},
  author={Zimmer, Walter and Cre{\ss}, Christian and Nguyen, Huu Tung and Knoll, Alois C},
  booktitle={2023 IEEE 26th International Conference on Intelligent Transportation Systems (ITSC)},
  pages={1030--1037},
  year={2023},
  organization={IEEE}
}

@inproceedings{xiang2024v2x,
  title={V2x-real: a largs-scale dataset for vehicle-to-everything cooperative perception},
  author={Xiang, Hao and Zheng, Zhaoliang and Xia, Xin and Xu, Runsheng and Gao, Letian and Zhou, Zewei and Han, Xu and Ji, Xinkai and Li, Mingxi and Meng, Zonglin and others},
  booktitle={European Conference on Computer Vision},
  pages={455--470},
  year={2024},
  organization={Springer}
}

@article{li2024choose,
  title={Choose your simulator wisely: A review on open-source simulators for autonomous driving},
  author={Li, Yueyuan and Yuan, Wei and Zhang, Songan and Yan, Weihao and Shen, Qiyuan and Wang, Chunxiang and Yang, Ming},
  journal={IEEE Transactions on Intelligent Vehicles},
  year={2024},
  publisher={IEEE}
}

@inproceedings{manivasagam2020lidarsim,
  title={Lidarsim: Realistic lidar simulation by leveraging the real world},
  author={Manivasagam, Sivabalan and Wang, Shenlong and Wong, Kelvin and Zeng, Wenyuan and Sazanovich, Mikita and Tan, Shuhan and Yang, Bin and Ma, Wei-Chiu and Urtasun, Raquel},
  booktitle={Proceedings of the IEEE/CVF Conference on Computer Vision and Pattern Recognition},
  pages={11167--11176},
  year={2020}
}

@inproceedings{dosovitskiy2017carla,
  title={CARLA: An open urban driving simulator},
  author={Dosovitskiy, Alexey and Ros, German and Codevilla, Felipe and Lopez, Antonio and Koltun, Vladlen},
  booktitle={Conference on robot learning},
  pages={1--16},
  year={2017},
  organization={PMLR}
}

@misc{team2020deepdrive,
  title={Deepdrive: a simulator that allows anyone with a pc to push the state-of-the-art in self-driving},
  author={Team, Deepdrive},
  year={2020}
}

@inproceedings{amini2022vista,
  title={Vista 2.0: An open, data-driven simulator for multimodal sensing and policy learning for autonomous vehicles},
  author={Amini, Alexander and Wang, Tsun-Hsuan and Gilitschenski, Igor and Schwarting, Wilko and Liu, Zhijian and Han, Song and Karaman, Sertac and Rus, Daniela},
  booktitle={2022 International Conference on Robotics and Automation (ICRA)},
  pages={2419--2426},
  year={2022},
  organization={IEEE}
}

@INPROCEEDINGS{8864642,
  author={Dworak, Daniel and Ciepiela, Filip and Derbisz, Jakub and Izzat, Izzat and Komorkiewicz, Mateusz and Wójcik, Mateusz},
  booktitle={2019 24th International Conference on Methods and Models in Automation and Robotics (MMAR)}, 
  title={Performance of LiDAR object detection deep learning architectures based on artificially generated point cloud data from CARLA simulator}, 
  year={2019},
  volume={},
  number={},
  pages={600-605},
  doi={10.1109/MMAR.2019.8864642}
}

@article{nowruzi2019much,
  title={How much real data do we actually need: Analyzing object detection performance using synthetic and real data},
  author={Nowruzi, Farzan Erlik and Kapoor, Prince and Kolhatkar, Dhanvin and Hassanat, Fahed Al and Laganiere, Robert and Rebut, Julien},
  journal={arXiv preprint arXiv:1907.07061},
  year={2019}
}

@misc{OpenStreetMap,
   author = {{OpenStreetMap contributors}},
   title = {{Planet dump retrieved from https://planet.osm.org }},
   howpublished = "\url{ https://www.openstreetmap.org }",
   year = {2017},
}

@inproceedings{zhao2021epointda,
  title={epointda: An end-to-end simulation-to-real domain adaptation framework for lidar point cloud segmentation},
  author={Zhao, Sicheng and Wang, Yezhen and Li, Bo and Wu, Bichen and Gao, Yang and Xu, Pengfei and Darrell, Trevor and Keutzer, Kurt},
  booktitle={Proceedings of the AAAI Conference on Artificial Intelligence},
  volume={35},
  number={4},
  pages={3500--3509},
  year={2021}
}

@article{li2025digital,
  title={Digital twin-assisted graph matching multi-task object detection method in complex traffic scenarios},
  author={Li, Mi and Liu, Chuhui and Pan, Xiaolong and Li, Zirui},
  journal={Scientific Reports},
  volume={15},
  number={1},
  pages={10847},
  year={2025},
  publisher={Nature Publishing Group UK London}
}

@inproceedings{strunz2024cross,
  title={Cross-Dataset Generalization: Bridging the Gap Between Real and Synthetic LiDAR Data},
  author={Strunz, Martin and Protzmann, Robert and Radusch, Ilja},
  booktitle={International Conference on Simulation Tools and Techniques},
  pages={202--217},
  year={2024},
  organization={Springer}
}

@data{ucf_ut_lumpi,
author = {Shahbaz, Muhammad and Agarwal, Shaurya},
publisher = {Harvard Dataverse},
title = {{UT-LUMPI}},
year = {2025},
version = {V1},
doi = {10.7910/DVN/D9SSWD},
url = {https://doi.org/10.7910/DVN/D9SSWD}
}

@data{ucf_ut_v2x_real_ic,
author = {Shahbaz, Muhammad and Agarwal, Shaurya},
publisher = {Harvard Dataverse},
title = {{UT-V2X-Real-IC}},
year = {2025},
version = {V2},
doi = {10.7910/DVN/N6N4UR},
url = {https://doi.org/10.7910/DVN/N6N4UR}
}

@data{ucf_ut_tumtraf_i,
author = {Shahbaz, Muhammad and Agarwal, Shaurya},
publisher = {Harvard Dataverse},
title = {{UT-TUMTraf-I}},
year = {2025},
version = {V2},
doi = {10.7910/DVN/D21HNZ},
url = {https://doi.org/10.7910/DVN/D21HNZ}
}

@inproceedings{Sun2020,
author = {Pei Sun and Henrik Kretzschmar and Xerxes Dotiwalla and Aur'{e}lien Chouard and Vijaysai Patnaik and Paul Tsui and James Guo and Yin Zhou and Yuning Chai and Benjamin Caine and Vijay Vasudevan and Wei Han and Jiquan Ngiam and Hang Zhao and Aleksei Timofeev and Scott Ettinger and Maxim Krivokon and Amy Gao and Aditya Joshi and Sheng Zhao and Shuyang Cheng and Yu Zhang and Jonathon Shlens and Zhifeng Chen and Dragomir Anguelov},
title = {{Scalability in Perception for Autonomous Driving: Waymo Open Dataset}},
booktitle = {Proc. IEEE/CVF Conference on Computer Vision and Pattern Recognition (CVPR) Workshops},
year = {2020}
}

@article{shahbaz2025,
author = {Shahbaz, Muhammad and Agarwal, Shaurya},
doi = {10.21105/joss.06751},
journal = {Journal of Open Source Software},
month = jun,
number = {110},
pages = {6751},
title = {{LiGuard: Interactively and Rapidly Create Point-Cloud and Image Processing Pipelines}},
url = {https://joss.theoj.org/papers/10.21105/joss.06751},
volume = {10},
year = {2025}
}

@inproceedings{carla,
	title        = {{CARLA}: {An} Open Urban Driving Simulator},
	author       = {Alexey Dosovitskiy and German Ros and Felipe Codevilla and Antonio Lopez and Vladlen Koltun},
	year         = 2017,
	booktitle    = {Proceedings of the 1st Annual Conference on Robot Learning},
	pages        = {1--16}
}

@misc{cesium_ion,
  author       = {{Cesium GS, Inc.}},
  title        = {{Cesium ion: 3D Geospatial Data Platform}},
  howpublished = {\url{https://cesium.com/platform/cesium-ion/}},
  note         = {Accessed: 2025-06-22},
  year         = {2025},
}

@misc{opentopography,
  author       = {{OpenTopography Community}},
  title        = {{OpenTopography: High‑Resolution Earth Science Data Portal}},
  howpublished = {\url{https://opentopography.org/}},
  note         = {Accessed: 2025‑06‑22},
  year         = {2025},
}

@software{autodesk_recappro,
  author       = {{Autodesk, Inc.}},
  title        = {{Autodesk ReCap Pro}},
  version      = {2026},
  url          = {https://www.autodesk.com/products/recap/overview},
  year         = {2025},
  note         = {Accessed: 2025‑06‑22},
}

@software{realitycapture,
  author       = {{Capturing Reality (Epic Games)}},
  title        = {{RealityCapture}},
  version      = {latest},
  url          = {https://www.capturingreality.com/},
  year         = {2025},
  note         = {Accessed: 2025‑06‑22},
}

@misc{colmap,
  author       = {{Johannes L. Schönberger and COLMAP developers}},
  title        = {{COLMAP: Structure‑from‑Motion and Multi‑View Stereo Software}},
  howpublished = {\url{https://colmap.github.io/}},
  note         = {Accessed: 2025‑06‑22},
  year         = {2025},
}

@software{autodesk_maya,
  author       = {{Autodesk, Inc.}},
  title        = {{Autodesk Maya}},
  version      = {2026},
  url          = {https://www.autodesk.com/products/maya/overview},
  year         = {2025},
  note         = {Accessed: 2025‑06‑22},
}

@software{cinema4d,
  author       = {{Maxon Computer GmbH}},
  title        = {{Cinema 4D}},
  version      = {2024},
  url          = {https://www.maxon.net/en/cinema‑4d},
  year         = {2025},
  note         = {Accessed: 2025‑06‑22},
}

@software{blender,
  author       = {{Stichting Blender Foundation and the Blender Community}},
  title        = {{Blender: 3D Modeling and Rendering Package}},
  version      = {latest},
  url          = {https://www.blender.org/},
  year         = {2025},
  note         = {Accessed: 2025‑06‑22},
}

@misc{google_maps,
  author       = {{Google LLC}},
  title        = {{Google Maps}},
  howpublished = {\url{https://maps.google.com/}},
  note         = {Accessed: 2025-06-22},
  year         = {2025},
}

@software{mathworks_roadrunner,
  author       = {{MathWorks, Inc.}},
  title        = {{RoadRunner: 3D Driving Scene Design Tool}},
  version      = {latest},
  url          = {https://www.mathworks.com/products/roadrunner.html},
  note         = {Accessed: 2025-06-22},
  year         = {2025}
}

@software{unreal_engine,
  author       = {{Epic Games}},
  title        = {{Unreal Engine}},
  version      = {4.26.2},
  date         = {2025-06-03},
  url          = {https://www.unrealengine.com/},
  note         = {Accessed: 2025-06-22},
}

@misc{shahbaz2025urbantwinhighfidelitysyntheticreplicas,
      title={UrbanTwin: High-Fidelity Synthetic Replicas of Roadside Lidar Datasets}, 
      author={Muhammad Shahbaz and Shaurya Agarwal},
      year={2025},
      eprint={2509.06781},
      archivePrefix={arXiv},
      primaryClass={cs.CV},
      url={https://arxiv.org/abs/2509.06781}, 
}

@misc{shahbaz2025highfidelitydigitaltwinsbridging,
      title={High-Fidelity Digital Twins for Bridging the Sim2Real Gap in LiDAR-Based ITS Perception}, 
      author={Muhammad Shahbaz and Shaurya Agarwal},
      year={2025},
      eprint={2509.02904},
      archivePrefix={arXiv},
      primaryClass={cs.CV},
      url={https://arxiv.org/abs/2509.02904}, 
}

@phdthesis{Unali24,
  author={Ozan Unal},
  title={Data-Efficient LiDAR Semantic Segmentation},
  year={2024},
  cdate={1704067200000},
  url={https://hdl.handle.net/20.500.11850/668173},
  school={ETH Zurich, Zürich, Switzerland}
}

\newpage

\begin{IEEEbiography}[{\includegraphics[width=1in,height=1.25in,clip,keepaspectratio]{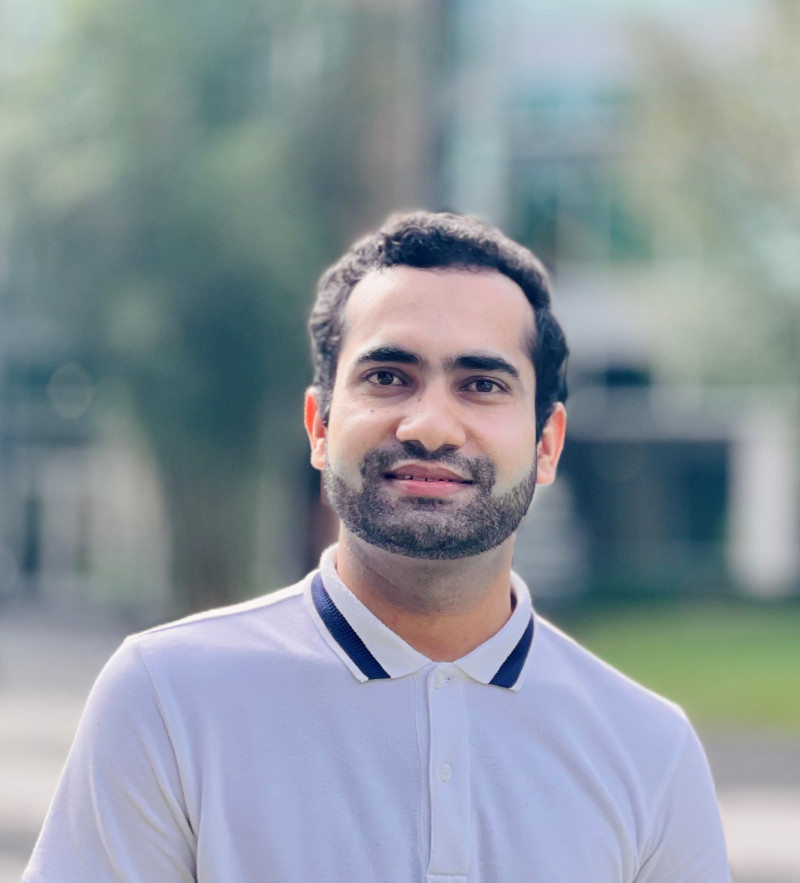}}]{Muhammad Shahbaz} is a post-doctoral scholar in Civil, Environmental and Construction Engineering department at University of Central Florida. He received the B.S. in computer science degree from Pir Mehr Ali Shah Arid Agriculture University Rawalpindi, and M.S. degree in computer science from the Pakistan Institute of Engineering and Applied Sciences, Islamabad, Pakistan. His research focuses on scalable LiDAR-based perception systems with a focus on Intelligent Transportation Systems (ITS). His work leverages self-supervised and Sim2Real learning techniques to develop efficient and reliable perception solutions. More generally, he is passionate about interdisciplinary research across Advanced Computer Vision and AI, and their applications in the field of volumetric perception and machine intelligence.
\end{IEEEbiography}

\begin{IEEEbiography}[{\includegraphics[width=1in,height=1.25in,clip,keepaspectratio]{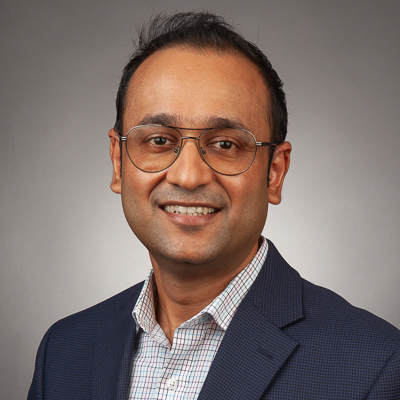}}]
{SHAURYA AGARWAL } 
 (Senior Member, IEEE) is currently an Associate Professor in the Civil, Environmental, and Construction Engineering Department at the University of Central Florida. He is the founding director of the Urban Intelligence and Smart City (URBANITY) Lab, Director of the Future City Initiative at UCF, and serves as the coordinator for Smart Cities Masters program at UCF. He was previously (2016-18) an Assistant Professor in the Electrical and Computer Engineering Department at California State University, Los Angeles. He completed his post-doctoral research at New York University (2016) and his Ph.D. in Electrical Engineering from the University of Nevada, Las Vegas (2015). His B.Tech. degree is in ECE from the Indian Institute of Technology (IIT), Guwahati. His research focuses on interdisciplinary areas of cyber-physical systems, smart and connected transportation, and connected and autonomous vehicles. Passionate about cross-disciplinary research, he integrates control theory, information science, data-driven techniques, and mathematical modeling in his work. As of May 2025, he has published a book, over 37 peer-reviewed journal publications, and multiple conference papers. His work has been funded by several private and government agencies. He is a \textit{senior member} of IEEE and serves as an \textit{Associate Editor} of IEEE Transactions on Intelligent Transportation Systems.
\end{IEEEbiography}

\vfill

\end{document}